\documentclass[conference]{IEEEtran}
\usepackage{caption}
\usepackage{graphicx}
\usepackage{amsmath, amssymb}
\usepackage{url}
\usepackage{listings}
\usepackage{color}
\usepackage{xcolor}
\usepackage{algorithm}
\usepackage{algorithmic}
\usepackage{amsmath,subcaption}
\usepackage{booktabs}  

\definecolor{codegray}{gray}{0.95}
\title{BakuFlow: A Streamlining Semi-Automatic Label Generation Tool}

\author{
\IEEEauthorblockN{Jerry Lin\IEEEauthorrefmark{1} and Patrick P. W. Chen\IEEEauthorrefmark{2}}
\IEEEauthorblockA{\IEEEauthorrefmark{1}BakuAI AS, Bergen, Norway \\}
\IEEEauthorblockA{\IEEEauthorrefmark{2}Silesian University of Technology, Poland \\
Email: jerry@bakuai.no; chenpiway@gmail.com}
}

\begin{document}

\maketitle

\begin{abstract}Accurately labeling (or annotation) data is still a bottleneck in computer vision, especially for large-scale tasks where manual labeling is time-consuming and error-prone. While tools like LabelImg can handle the labeling task, some of them still require annotators to manually label each image. In this paper, we introduce BakuFlow, a streamlining semi-automatic label generation tool. Key features include (1) a live adjustable magnifier for pixel-precise manual corrections, improving user experience; (2) an interactive data augmentation module to diversify training datasets; (3) label propagation for rapidly copying labeled objects between consecutive frames, greatly accelerating annotation of video data; and (4) an automatic labeling module powered by a modified YOLOE framework. Unlike the original YOLOE, our extension supports adding new object classes and any number of visual prompts per class during annotation, enabling flexible and scalable labeling for dynamic, real-world datasets. These innovations make BakuFlow especially effective for object detection and tracking, substantially reducing labeling workload and improving efficiency in practical computer vision and industrial scenarios.

\end{abstract}


\section{Introduction}
Training an AI-powered system with accurate object detection~
\cite{ref6,ref7,ref8,ref9,ref10} and recognition is an important task~\cite{ref1,ref2}, especially for visually oriented applications. You Only Look Once (YOLO)-based series models~\cite{ref11,ref12,ref13,ref14} are the standard approaches for object detection, and many existing tools, such as LabelImg\footnote{https://github.com/HumanSignal/labelImg}, CVAT\footnote{https://github.com/cvat-ai/cvat}, LabelMe\footnote{https://github.com/LabelMe/labelme}, Roboflow Annotate\footnote{https://roboflow.com/annotate}, VoTT\footnote{https://github.com/microsoft/VoTT}, and Label Studio\footnote{https://github.com/HumanSignal/label-studio} support the labeling of data by allowing the user to draw bounding boxes (bboxes) on images for different scenarios (e.g., detection). However, these tools are not well-suited for efficient processing of sequential or consecutive images extracted from video clips. With such tools, all objects in an image must be manually labeled, which usually takes 1–3 seconds per object. This becomes extremely inefficient when an image contains more than hundreds or thousands of objects with a manual labeling process. Several studies~\cite{ref4,ref5} focused on using AI/ML/DL for auto-labeling, but in many real-world scenarios, e.g.,
 based on our practical experience in retail, video images are processed as sequences of consecutive frames, with some objects being moved or bought between frames (e.g. planogram compliance application). Although some labeling tools (e.g., CVAT, Label Studio, Roboflow Annotate) support an auto-labeling function, they are not designed with a simplified, portable, and user-friendly UI and the learning curve by using those tools is long. Therefore, it is essential to develop a simplified and efficient labeling tool that is tailored to process such sequential and consecutive images.


We propose \textbf{BakuFlow}\footnote{https://github.com/bakuai-as/BakuFlow}$^{,}$\footnote{https://youtu.be/QJZYhl6e6HM}, an AI-powered labeling tool that leverages state-of-the-art visual prompting techniques, with YOLOE~\cite{ref3} as the core of its auto-labeling module for both single-image and cross-image labeling. 
Unlike traditional labeling tools, BakuFlow leverages YOLOE’s visual prompt embeddings to automate labeling for sequential images, substantially reducing manual workload. The original YOLOE only supports a single visual prompt for all classes, limiting its ability to generalize to diverse object appearances. In contrast, our implementation enables users to assign multiple visual prompts to each class, capturing greater intra-class variation and improving annotation accuracy and robustness for complex or evolving datasets.
This capability enables efficient, scalable labeling workflows for real-world datasets that evolve over time.
Beyond its advanced auto-labeling core, BakuFlow addresses key limitations of existing annotation tools by introducing efficient label propagation across sequential images. Users can easily copy selected or all labels from one frame to the next, significantly streamlining the labeling of video frames and burst images. BakuFlow also features an adjustable magnifier view for precise bounding box adjustment. Additional user-centric functionalities, such as undo/redo, auto-save, and dynamic language recognition can further enhance workflow efficiency and minimize annotation errors.

In summary, BakuFlow offers:
\begin{enumerate}
    \item Efficient labels propagation and management for sequential images;
    \item An adjustable live magnifier for high-precision labeling and correction;
    \item Data augmentation module enhances training effectiveness by generating diverse image variations, enabling robust model scalability with rich training data;
    \item A YOLOE-based~\cite{ref3} auto-labeling engine that, unlike the original version which only allows a single prompt for all classes, supports flexible class management and multiple prompts for each class.
\end{enumerate}


\section{System Architecture}

BakuFlow is a robust, cross-platform annotation tool developed in Python using PyQt5 for the user interface and OpenCV\footnote{https://opencv.org/} for efficient image labeling. The system is designed with a modular architecture comprising:

\begin{itemize}
\item \textbf{GUI Core}: A responsive and portable interface built with Qt widgets for efficient image display, navigation, and direct user interaction.
\item \textbf{Advanced Annotation Engine}: Supports creation, editing, and multi-selection of bounding boxes, with advanced module (e.g., labels propagation) for inheritance and bulk operations across sequential frames. In addition, an interactive magnification module supports precise annotation by enabling real-time zoom and fine-tuned bbox adjustments for accurate object labeling.
\item \textbf{Data I/O}: Handles all major image formats (e.g., JPG, PNG, WebP) and exports to standard annotation schemas (YOLO, VOC, COCO), ensuring compatibility with common computer vision pipelines.
\item \textbf{Localization}: Offers multi-language support and automatic language detection, making the tool accessible to international teams.
\item \textbf{Data Augmentation}:
Offers interactive data augmentation tools, including brightness, contrast, rotation, and flipping. It enables users to instantly diversify their annotated datasets. This feature helps improve the generalizability and robustness of AI models trained on the labeled data, especially when working with limited or imbalanced real-world samples.
\item \textbf{Auto-Labeling}:
Features a YOLOE-based~\cite{ref3} auto-labeling module that overcomes the original framework’s limitation of supporting only a single visual prompt per class. In BakuFlow, users can assign multiple prompts to each class as needed, providing greater flexibility and accuracy in labeling complex or evolving datasets.
\end{itemize}


\begin{figure*}[!htbp]
    \centering
    \begin{subfigure}{0.25\textwidth}
        \centering
        \includegraphics[width=\linewidth]{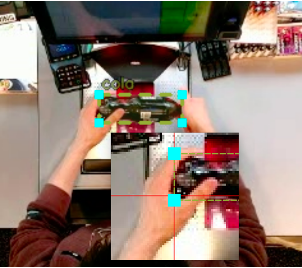}
        \caption{A live magnifier}
        \label{fig1}
    \end{subfigure}
    \hfill
    \begin{subfigure}{0.25\textwidth}
        \centering
        \includegraphics[width=0.8\linewidth]{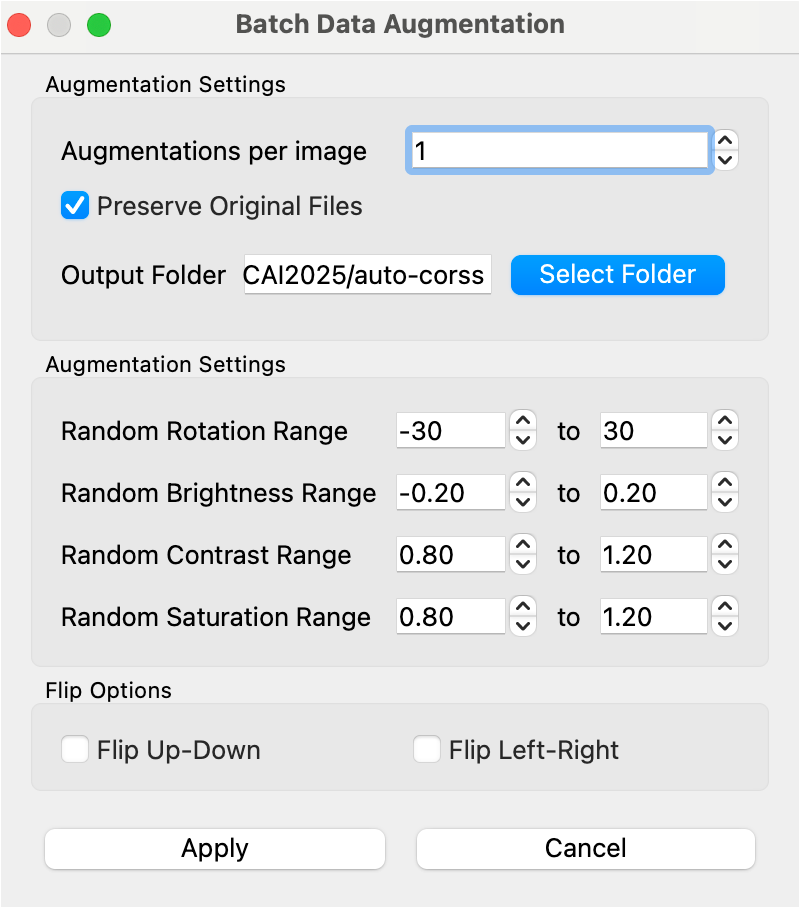}
        \caption{Data augmentation}
        \label{fig2}
    \end{subfigure}
    \hfill
    \begin{subfigure}{0.4\textwidth}
        \centering
        \includegraphics[width=\linewidth]{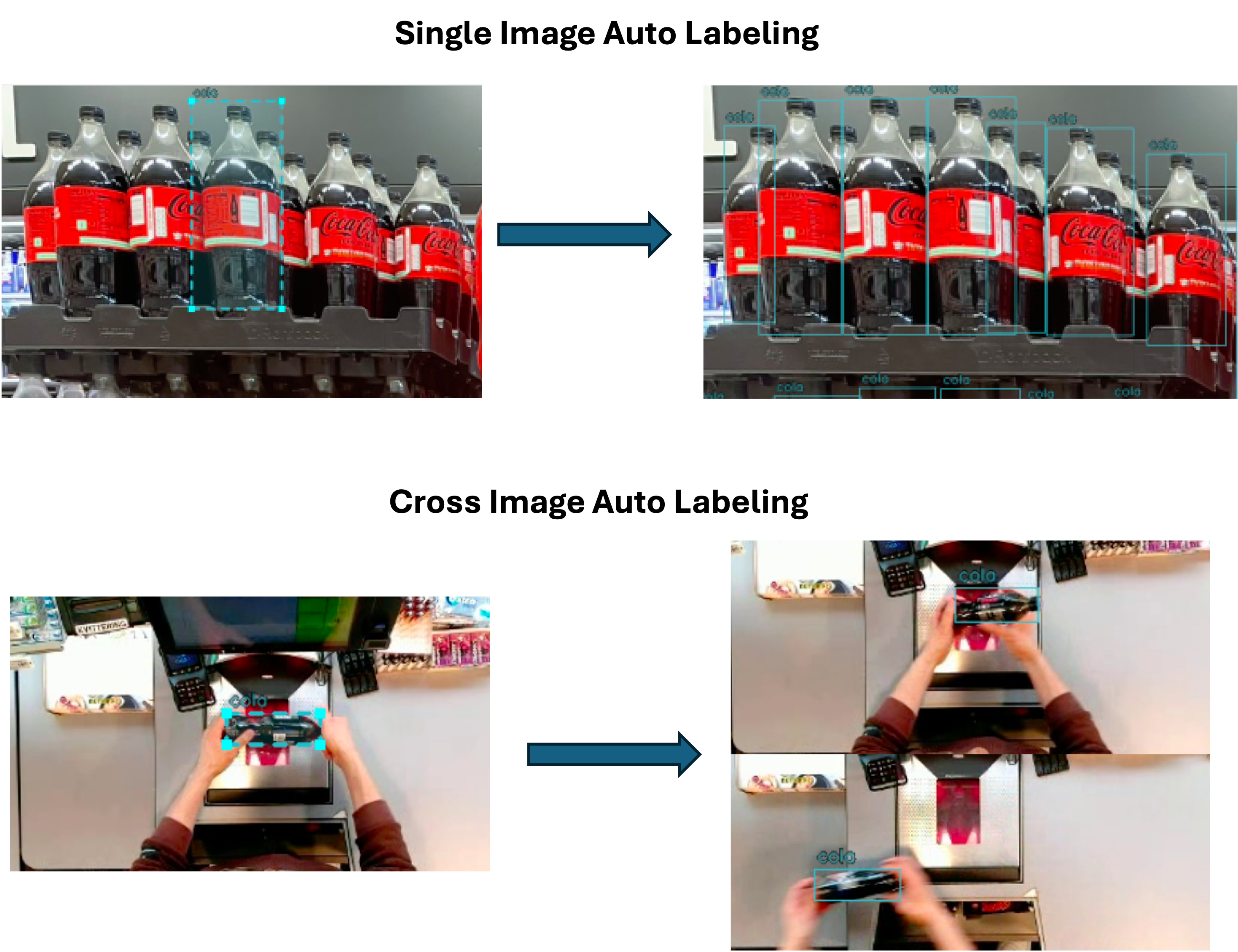}
        \caption{Auto-labeling}
        \label{fig3}
    \end{subfigure}
    \caption{Overview of major features: (a) Live magnifier; (b) Data augmentation; (c) Auto-labeling}
    \label{fig:main}
\end{figure*}

\section{Key Functionalities}
To further reduce annotation errors and accidental loss, BakuFlow supports undo/redo actions and automatic saving, ensuring smooth and reliable operation throughout long labeling sessions. Enhanced labeling efficiency through intuitive keyboard shortcuts, enabling rapid annotation and seamless workflow control.
%
%
Major functions implemented in BakuFlow are stated as follows. 

\subsection{Optimized Annotation Workflow and User Experience}

BakuFlow is designed with a strong emphasis on user efficiency and annotation precision, featuring a range of interactive tools that streamline the labeling process:

\begin{itemize}
\item \textbf{Intuitive Bbox Editing:} Users can quickly draw, move, and resize bounding boxes using intuitive mouse gestures. To enhance clarity in dense scenes, bboxes for different classes are automatically color-coded for easy visual distinction.
\item \textbf{Multi-Selection and Label Propagation:} Annotators can duplicate all or selected bounding boxes from one frame to the next, a feature particularly effective for sequential video frames with minor object movement. This capability reduces repetitive work and improves consistency.
\item \textbf{Live Magnifier View:} For challenging images such as those with low resolution or densely packed objects, an adjustable live magnifier follows the cursor during bounding box operations, enabling pixel-level accuracy and fine-grained control. The magnifier’s size is fully adjustable to accommodate user preference, which can be seen in  Fig.~\ref{fig:main}(\subref{fig1}).

\end{itemize}

\subsection{Data Augmentation}
Data augmentation\footnote{https://docs.ultralytics.com/guides/yolo-data-augmentation/} is a widely used technique in computer vision in which the training dataset is artificially extended by applying various transformations to existing images. This approach increases data diversity and helps to improve the robustness of AI-powered models by reducing overfitting and improving generalization under different environmental conditions. To support this, the developed software includes configurable parameters for \textbf{rotation}, \textbf{brightness}, \textbf{contrast}, \textbf{saturation}, \textbf{vertical flipping (top to bottom)}, \textbf{horizontal flipping (leftto right)} and the \textbf{number of enriched samples} per image. These settings allow users to create enriched training datasets tailored to specific use cases. An UI for data augmentation is shown in Fig.~\ref{fig:main}(\subref{fig2}).

\subsection{Auto-Labeling with Flexible Visual Prompts}

At the core of BakuFlow is an auto-labeling module powered by YOLOE~\cite{ref3}, which leverages visual prompt embeddings (VPEs) to guide object detection. Our motivation stems from the observation that providing a greater number and diversity of visual prompts for each object class leads to more robust and accurate auto-labeling, especially in varied or challenging datasets. By capturing a wider range of intra-class variations through multiple prompts, the detection model can better generalize to real-world data.

However, the original YOLOE framework is limited to a single visual prompt for all classes, restricting its ability to represent diverse object appearances and limiting its flexibility in real-world annotation tasks. BakuFlow overcomes this limitation by allowing users to assign multiple, varied visual prompts to each class, thus enhancing the robustness and adaptability of the auto-labeling process.

This flexible prompt management enables two major auto-labeling scenarios:
\begin{itemize}
\item \textbf{Single-Image Auto-Labeling:} BakuFlow can automatically annotate multiple objects within a single image using visual prompts, significantly accelerating the labeling of complex scenes.
\item \textbf{Cross-Image Auto-Labeling:} The system supports automated object annotation across consecutive images or video frames, leveraging consistent visual prompts to efficiently propagate labels throughout sequential data.
\end{itemize}

An example of the user interface for managing prompts and auto-labeling is shown in Fig.~\ref{fig:main}(\subref{fig3}).


In summary, BakuFlow stands out with the following key features:
\begin{itemize}
    \item \textbf{Comprehensive user experience:} Live magnifier for pixel-level accuracy, undo/redo, auto-save, multi-language support, and a lightweight, portable design, making BakuFlow accessible and practical for global users;

    \item \textbf{Optimized annotation workflow:} Efficient propagation of labels, single (multi)-selection support, duplication of annotations, and interactive data augmentation support for a clear and responsive user interface;

    \item \textbf{Flexible and robust auto-labeling core:} An enhanced YOLOE-based auto-labeling engine that supports multiple classes and multiple visual prompts per class, allowing the model to capture greater object variability and achieve more accurate, generalizable labeling results in real-world scenarios.
\end{itemize}

\section{Comparison and Evaluation in Use Cases}

We compare BakuFlow with two widely used annotation tools, LabelImg and CVAT, across key aspects relevant to modern image labeling workflows.

\subsection{Functional Comparison}
A summary of practical features and workflow efficiency is presented in Table~\ref{tab:comparison}. BakuFlow stands out for its integrated, dynamic auto-labeling and optimized user experience, while LabelImg and CVAT have different strengths in deployment and extensibility.

\begin{table}[!htbp]
    \centering
    \caption{Feature comparison between BakuFlow, LabelImg, and CVAT.}
    \label{tab:comparison}
    \begin{tabular}{@{} lccc @{}}
        \toprule
        \textbf{Feature} & \textbf{BakuFlow} & \textbf{LabelImg} & \textbf{CVAT} \\
        \midrule
        Auto-labeling (YOLOE)       & \checkmark~(dynamic) & $\times$ & \checkmark~(static) \\
        Label propagation           & \checkmark~(bulk/sel.) & $\times$ & \checkmark~(limited) \\
        Data augmentation           & \checkmark~(UI) & $\times$ & $\times$~(plugins) \\
        Magnifier                   & \checkmark & $\times$ & $\times$ \\
        Multi-language UI           & \checkmark & \checkmark & \checkmark \\
        Undo/Redo, Auto-save        & \checkmark & $\times$ & \checkmark \\
        Local/Portable              & \checkmark & \checkmark & $\times$~(server) \\
        \bottomrule
    \end{tabular}
\end{table}

\section{Future Work and Conclusion}
BakuFlow delivers a novel image labeling solution, integrating label propagation, an adjustable live magnifier, data augmentation, and flexible visual prompt management for efficient labeling in both single and cross images scene. The current platform demonstrates improved usability and labeling accuracy for real-world computer vision applications.

For future work, we plan several directions to further enhance BakuFlow. First, our empirical observations show that as the number of classes to be detected from a single visual prompt increases, the stability and accuracy of YOLOE-based detection may degrade. This reveals a limitation in the current YOLOE mechanism, which we aim to address in future iterations by exploring better improved prompt representations, more effective prompt-class assignment strategies, and possible architectural modifications. Additionally, we plan to investigate model optimization and quantization, collaborative annotation functionality, and extended annotation support for other YOLO-based models or architectures. An AI visual platform will be integrated with BakuFlow and explored be for better labeling, training, and deployment with one-click function.



\bibliographystyle{IEEEtran}

\end{document}